\documentclass[letterpaper]{article} 
\usepackage{aaai25}  
\usepackage{times}  
\usepackage{helvet}  
\usepackage{courier}  
\usepackage[hyphens]{url}  
\usepackage{graphicx} 
\usepackage{natbib}  
\usepackage{caption} 
\usepackage{algorithm}
\usepackage{algorithmic}
\usepackage{tabularray}
\usepackage{amsmath}
\usepackage{amssymb} 
\newcommand{\ra}[1]{\renewcommand{\arraystretch}{#1}}
\usepackage{newfloat}
\usepackage{listings}
\DeclareCaptionStyle{ruled}{labelfont=normalfont,labelsep=colon,strut=off} 
\floatstyle{ruled}

\newfloat{listing}{tb}{lst}{}
\floatname{listing}{Listing}
\title{Causal-Inspired Multitask Learning for Video-Based Human Pose Estimation}
\author{
    Haipeng Chen\textsuperscript{\rm 1,2},
    Sifan Wu\textsuperscript{\rm 1,2}\thanks{Corresponding to: Yingying Jiao, Sifan Wu, Zhigang Wang, Yingda Lyu.},
    Zhigang Wang\textsuperscript{\rm 3}\footnotemark[1],\\
    Yifang Yin\textsuperscript{\rm 4},
    Yingying Jiao\textsuperscript{\rm 1,2}\footnotemark[1],
    Yingda Lyu\textsuperscript{\rm 1,5}\footnotemark[1],
    Zhenguang Liu\textsuperscript{\rm 6,7}
}
\affiliations{
    \textsuperscript{\rm 1}College of Computer Science and Technology, Jilin University,\\
    \textsuperscript{\rm 2}Key Laboratory of Symbolic Computation and Knowledge Engineering of Ministry of Education, Jilin University,\\
    \textsuperscript{\rm 3}College of Computer Science and Technology, Zhejiang Gongshang University,\\
    \textsuperscript{\rm 4}Institute for Infocomm Research ($I^{2}R$), A*STAR,\\
    \textsuperscript{\rm 5}Public Computer Education and Research Center, Jilin University,\\
    \textsuperscript{\rm 6}The State Key Laboratory of Blockchain and Data Security, Zhejiang University,\\
    \textsuperscript{\rm 7}Hangzhou High-Tech Zone (Binjiang) Institute of Blockchain and Data Security,\\
    chenhp@jlu.edu.cn, wusifan2021@gmail.com, wangzhigang2024@gmail.com,\\ yin\_yifang@i2r.a-star.edu.sg, jiaoyy21@mails.jlu.edu.cn, ydlv@jlu.edu.cn, liuzhenguang2008@gmail.com

}

\usepackage{bibentry}

\begin{document}

\maketitle

\begin{abstract}
Video-based human pose estimation has long been a fundamental yet challenging problem in computer vision. Previous studies focus on spatio-temporal modeling through the enhancement of architecture design and optimization strategies. However, they overlook the causal relationships in the joints, leading to models that may be overly tailored and thus estimate poorly to challenging scenes. Therefore, adequate causal reasoning capability, coupled with good interpretability of model, are both indispensable and prerequisite for achieving reliable results. In this paper, we pioneer a causal perspective on pose estimation and introduce a causal-inspired multitask learning framework, consisting of two stages. \textit{In the first stage}, we try to endow the model with causal spatio-temporal modeling ability by introducing two self-supervision auxiliary tasks. Specifically, these auxiliary tasks enable the network to infer challenging keypoints based on observed keypoint information, thereby imbuing causal reasoning capabilities into the model and making it robust to challenging scenes. \textit{In the second stage}, we argue that not all feature tokens contribute equally to pose estimation. Prioritizing causal (keypoint-relevant) tokens is crucial to achieve reliable results, which could improve the interpretability of the model. To this end, we propose a Token Causal Importance Selection module to identify the causal tokens and non-causal tokens (\textit{e.g.}, background and objects). Additionally, non-causal tokens could provide potentially beneficial cues but may be redundant. We further introduce a non-causal tokens clustering module to merge the similar non-causal tokens. Extensive experiments show that our method outperforms state-of-the-art methods on three large-scale benchmark datasets. 
\end{abstract}
\section{Introduction}
Estimating human poses from videos is a fundamental topic in artificial intelligence, which aims to identify and localize anatomical keypoints on the human body.
In recent years, this task has garnered an increasing interest from researchers and industry, accelerating advances in human-centric applications in diverse scenes \textit{such as} action recognition ~\cite{yang2023action}, motion prediction~\cite{su2021motion} and motion transfer~\cite{liu2022copy,wu2024pose}.


With the continual breakthroughs of deep learning algorithms and models~\cite{yao2024swift,RAMMVC}, AI has achieved success in various fields~\cite{shuai2023locate,chen2024pioneeringneuralnetworkmethod,SSGCC}. For pose estimation, one line of work focuses on designing different network structures. \cite{artacho2020unipose} adopts CNNs and LSTM to extract the intrinsic motion dynamic of persons. \cite{feng2023mutual} proposes feature difference method to capture spatio-temporal dependencies in videos. Another line of work introduces specific loss functions to supervise the network. \cite{liu2022temporal,feng2023mutual} propose a mutual information objective to align the features of sequences and obtain an informative representation.

Though promising, there are a few clouds on the horizon for video-based pose estimation. 1) \textbf{Robustness}. Over-tailored network structures~\cite{tang2024hunting,xu2024polyp} and the lack of causal perceptual capabilities compromise the robustness of models~\cite{zhang2023knowledge}. Challenging scenes such as pose occlusion and video defocus often appear in videos, where the model fails to accurately estimate the human poses. For instance, when multiple people play football, the legs of the persons in the video occlude each other. Existing methods primarily concentrate on the occluded parts and attempt to identify the positions of the occluded keypoints. However, they usually sacrifice an explicit understanding of observed (non-entangled region) visual cues. In other words, they lack an ability to causal reasoning, utilizing observed variables to speculate about unknown (entangled region) variables. 2) \textbf{Interpretability}. For pose estimation tasks, each frame is a mixture of causal (keypoint-relevant) and non-causal (\textit{e.g.,} background, objects) factors~\cite{liu2024causality}. Obviously, the contribution of causal factors is much greater than that of non-causal factors. Existing methods struggle to capture the causal features that are crucial for identifying the human pose, which deprives the network of model interpretability.



In this work, adopting a causal look at pose estimation, we present a  \underline{\textbf{C}}ausal-inspired \underline{\textbf{M}}ultitask learning \underline{\textbf{P}ose} estimation framework (CM-Pose) in a two-stage process. \textit{In the first stage}, we aim to foster the network with causal reasoning ability, enabling it to infer the locations of challenging (occluded or blurred) keypoints based on the observed keypoint information. Specifically, we present a novel multitask learning framework, which introduces different \textit{auxiliary tasks} on top of the \textit{primary pose estimation task}. Central to the idea of auxiliary tasks is to randomly corrupt partial feature tokens and then recover them utilizing normal tokens based on the causal spatio-temporal modeling capability of network. 
Our framework complements existing pose estimation approaches by emphasizing an effective multitask learning paradigm in a causal view.

Technically, for the challenging scenes of pose occlusion and video defocus, we propose two self-supervision auxiliary tasks: a masked token reconstruction task and a denoising token task. The masked token reconstruction task randomly masks the feature tokens, with the objective of reconstructing the masked tokens. This task simulates the pose occlusion scene where some keypoints are occluded. Similarly, the token denoising task randomly adds Gaussian noise to the feature tokens, with the goal of recovering the corrupted tokens. The task aims to simulate the video defocus scenario where some areas of the image become blurred. Compared with the existing masked/denoising learning paradigms, they always deal with data types such as images~\cite{he2022masked} and motion sequences~\cite{xu2023auxiliary}. To the best of our knowledge, we are the first work to propose masked/denoising tokens reconstruction learning to promote the primary pose estimation task. To better cultivate the spatio-temporal causal reasoning ability of multitask network, we introduce a simple yet effective network, a criss-cross spatio-temporal attention network. 
These tasks share the same network, and the auxiliary tasks are only employed during the training process, incurring no additional computational costs.

\textit{In the second stage}, we inject interpretability into our model. Not all feature tokens contribute equally to pose estimation. We argue that revealing ``which features of images is important for keypoint position" is the key to accurately and reliably estimate human poses. Specifically, we propose a Token Causal Importance Selection module and a Non-causal Tokens Clustering module. Firstly, given coarse tokens from multitask coarse learning, the token causal importance selection module differentiates these tokens into causal (keypoint region) and non-causal (background or objects) tokens based on keypoint tokens attention. While non-causal tokens are not directly related to human pose, they could supply the potential clues and improve the expressivity of the model~\cite{long2023beyond}. Therefore, it is inappropriate to directly discard non-causal tokens. However, non-causal tokens may contain redundant information, such as multiple background tokens with the same semantics. To this end, the non-causal tokens clustering module applies a density peaks clustering algorithm based on $k$ nearest neighbor to effectively cluster similar non-causal tokens and merge these tokens from the same group into a new token. Finally, causal tokens and merged non-causal tokens are aggregated to informative tokens, which are fed into a pose detection head to estimate the human pose. We believe that these insights will open avenues for future research on video-based human pose estimation.


\textbf{Contributions.} The key contributions of this work are summarized as follows:

\begin{itemize}
	\item[$\bullet$] In this paper, we investigate the video-based pose estimation task from a causal perspective, aiming to inject good robustness and interpretability into our model in challenging scenes.
	\item[$\bullet$] We propose a causal-inspired multitask learning framework that enables the model to perform causal spatio-temporal modeling on challenging scenes. We further introduce a token causal importance selection module and a non-causal token clustering module to identify causal features and compact (less redundancy) non-causal features, which improves the interpretability of the model.
	
	\item[$\bullet$] Our method achieves state-of-the-art performance on three benchmark datasets, \textit{i.e.,} PoseTrack2017, PoseTrack2018, PoseTrack2021.
\end{itemize}

\section{Related Works}
\textbf{Vision Transformer on Image-based Pose Estimation.} Vision transformer (ViT)~\cite{dosovitskiy2020image} provides an alternative to CNN-based methods for various visual tasks~\cite{tang2023duat,chen2023self,chen2024tokenunify}. \cite{li2021tokenpose} proposes Tokenpose, which first utilizes CNN to extract feature maps and then employs ViT to estimate the human pose in images. \cite{xu2022vitpose} first proposes to use ViT model without CNN to perform image-level pose estimation. Image-based pose estimation methods have gained perfect performance. However, they are not suitable for video-based pose estimation tasks because they focus only on intra-frame spatial relationships and ignore the abundant inter-frame temporal dependencies. 

\begin{figure*}[t]
	\centering
	\includegraphics[width=1\textwidth]{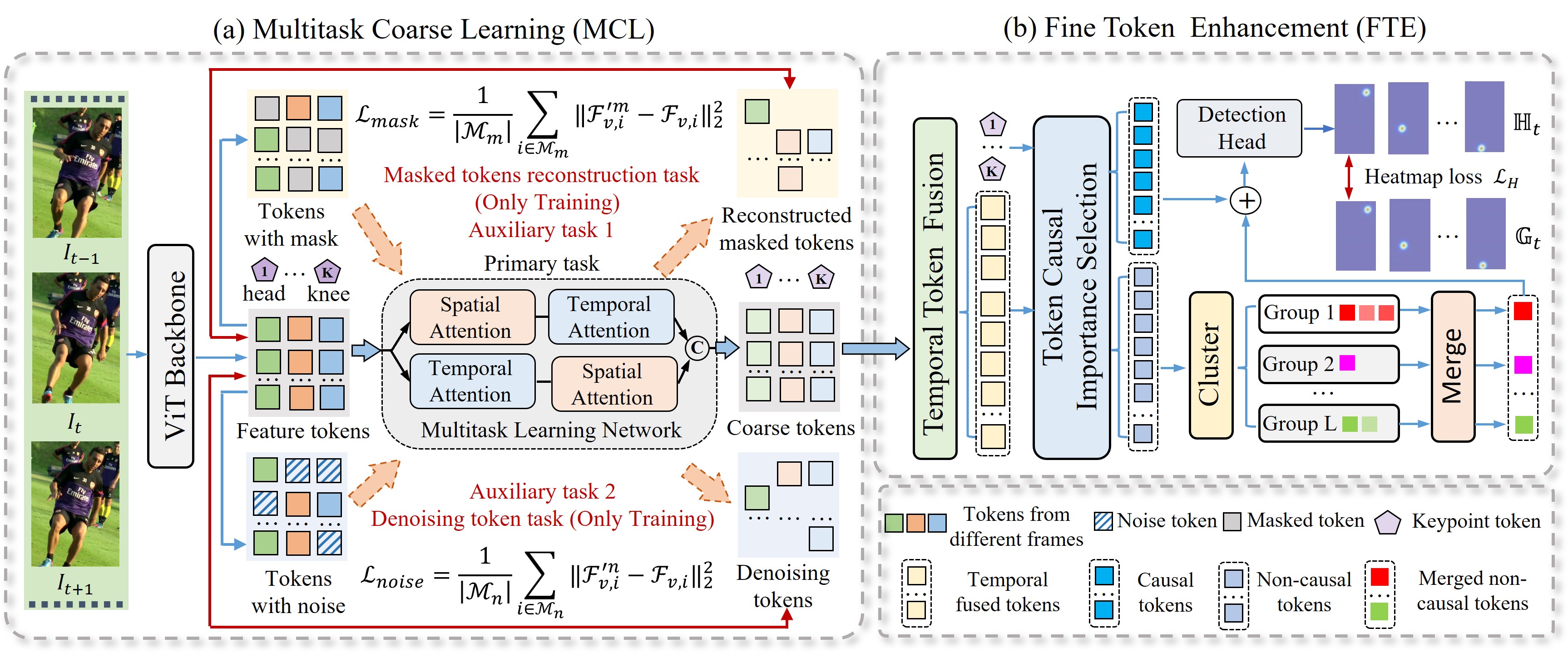}
	\caption{Overall pipeline of our CM-Pose framework. The goal is to identify the pose of keyframe $I_t$. CM-Pose includes two key components: Multitask Coarse Learning (MCL) and Fine Token Enhancement (FTE). (a) The MCL module consists of three tasks: the primary pose estimation task (middle branch), the masked token reconstruction task (upper branch), and the denoising token task (lower branch). To model the causal spatio-temporal dependencies for these tasks, we design a simple yet effective multitask learning network, a criss-cross spatio-temporal attention network. (b) The FTE module first performs temporal fusion on the coarse tokens. We then decouple the causal and non-causal tokens according to keypoint tokens attention. We further cluster non-causal tokens and merge the tokens from the same group into a new token. Finally, we concatenate the causal tokens and the merged non-causal tokens to estimate the pose $\mathbb{H}_t$ through a detection head.}
	\label{method}
\end{figure*}


\textbf{Video-Based Human Pose Estimation.} Existing methods~\cite{jin2022otpose,jiao2022glpose,wu2024joint} resort to different frameworks or loss functions to model spatio-temporal dependencies of videos. \cite{song2017thin} employ dense optical flow between frames to polish the pose of keyframe. \cite{feng2023mutual} introduces feature difference to capture spatial features and temporal dynamics. In addition to designing various network structures, \cite{liu2022temporal} proposes a mutual information loss function to supervise the network. In general, the performance of pose estimation is heavily dependent on the dedicated networks or loss functions. In contrast, we adopt a new training strategy that efficiently trains the network by leveraging causal-inspired multitask learning framework.

\textbf{Auxiliary Task Learning.} Auxiliary task learning is a novel learning paradigm that aims to simulate challenging scenes to make network more robust, imbuing the network with robust information. Auxiliary task learning has demonstrated extraordinary results in various tasks, such as video captioning \cite{gao2021hierarchical} and motion prediction \cite{xu2023auxiliary}. To the best of our knowledge, auxiliary task learning has not been attempted in pose estimation. In this paper, we propose CM-Pose, which explores the potential of auxiliary task learning in video-based human pose estimation.

\section{Method}


\subsection{Problem Formulation} Given a video sequence, the video-based human pose estimation task aims to identify the pose of all person in every frame. Technically, we first utilize an object detector~\cite{qiao2021detectors} to obtain the bounding boxes of each person in video frames. To crop the same person in the video sequence, we then enlarge all bounding boxes by 25\% and obtain the video clip of person $i$, \textit{i.e.}, $\mathcal{I}_t^i=\{I_{t-\omega}^i,...,I_t^i,...,I_{t+\omega}^i\}$ (where $\omega$ being a predefined time span) centered on the keyframe $I_t^i$. We are interested in modeling the spatio-temporal dependencies in $\mathcal{I}_t^i$ to estimate the pose in $I_t^i$. To facilitate representation and understanding, we set $\omega=1$ and omit the superscript $i$.

\subsection{Method Overview} As shown in Figure~\ref{method}, the proposed method CM-Pose comprises two key components: Multitask Coarse Learning (MCL) and Fine Token Enhancement (FTE). (1) We take a causal view to propose MCL, which learns the primary pose estimation task along with extra-designed auxiliary tasks. MCL enable the network to have more comprehensive causal spatio-temporal modeling capabilities and be robust to challenging scenes. 
(2) The FTE module categorizes the coarse tokens (generated from primary task in the MCL) into causal and non-causal tokens based on keypoint token attention scores~\cite{CMSCGC}, injecting good interpretability into the model.
In order to reduce the information redundancy of non-causal tokens, we cluster and merge similar non-causal tokens. Finally, we concatenate the causal tokens and the merged non-causal tokens to jointly infer human pose. In the following, we introduce the two components in detail.

\subsection{Multitask Coarse Learning} The key to accurately estimating pose is to model the spatial and temporal dependencies of joints based on feature tokens. Existing methods focus on spatio-temporal modeling through the enhancement of architecture design or optimization strategies. However, they ignore the causal correlation in the feature tokens, resulting in models that may be overly tailored and thus estimate poorly to challenging scenes. We consider a novel and interesting research line, a causal-inspired multitask learning framework, that imposes the network to capture more comprehensive causal spatio-temporal dependencies by introducing extra-designed auxiliary tasks. There are three important steps: feature extraction, auxiliary tasks, and multitask learning network.

\textbf{Feature Extraction.} Given a frame sequence $\mathcal{I}_t =\{ I_{t-1},I_t,I_{t+1}\}$, we first utilize a backbone network~\cite{dosovitskiy2020image} to generate the initial feature tokens $\{F_{t-1},F_t,F_{t+1}\}$. We then concatenate them to obtain $\mathcal{F}_v=F_{t-1}\oplus F_t\oplus F_{t+1}$, $\mathcal{F}_v\in \mathbb{R}^{3 N\times D}$, where $\oplus$ represents the concatenate operation, $N$ denotes the number of tokens in a frame and $D$ is the dimension of hidden embedding. Besides, following~\cite{li2021tokenpose}, we also introduce $K$ additional learnable keypoint tokens $\mathcal{F}_k\in  \mathbb{R}^{K\times D }$ to represent $K$ keypoints.

\textbf{Auxiliary Tasks.} To enable the network to perform robust pose estimation in challenging scenes (\textit{i.e.}, pose occlusion and video defocus), we propose two auxiliary tasks, respectively: masked token reconstruction task and denoising token task. Specifically, given initial tokens $\mathcal{F}_v$, in the masked token reconstruction task, each token of $\mathcal{F}_v$ is masked to zero value with a probability $p_{tm}$ (\textit{i.e.}, $3\cdot N \cdot p_{tm}$ tokens are masked), obtaining masked tokens $\mathcal{F}_v^m$. Similarly, in the denoising token task, each token of $\mathcal{F}_v$ has a probability $p_{tn}$ of being added with a Gaussian noise (\textit{i.e.}, $3\cdot N \cdot p_{tn}$ tokens are added noise). 
We can obtain the tokens with noise $\mathcal{F}_v^n$. 
The auxiliary task requires the network to reconstruct corrupted tokens by utilizing the limited semantic context available from normal tokens, which cultivates the causal spatio-temporal modeling ability of network and perform better to challenging scenes.
It is worth noting that introducing additional auxiliary tasks in the learning framework, which share the multitask learning network with primary task, does not increase the model size. During inference, we remove the auxiliary tasks. Besides, the flag of the masked tokens is formulated as: 
\begin{equation}
	\begin{aligned}
		M_m(i) = \begin{cases}
			0, & \text{if token } i \text{ is masked}, \\
			1, & \text{otherwise}.
		\end{cases}
	\end{aligned}
\end{equation}

Similarly, the flag of token denoising task is $M_n$, which helps in auxiliary tasks to supervise the network training utilizing reconstruction losses. Mathematically, assuming the masked token set $\mathcal{M}_m=\{i|M_m(i)=0\}$ and the noised token set $\mathcal{M}_n=\{i|M_n(i)=0\}$, the loss functions of two auxiliary tasks are formulated as:
\begin{align}
	\mathcal{L}_{mask} = \frac{1}{\left| \mathcal{M}_m\right|}\sum_{i\in \mathcal{M}_m} \left\|\mathcal{F}_{v,i}^{'m}-\mathcal{F}_{v,i}\right\|_2^2,
	\label{mask}\\
	\mathcal{L}_{denoise} = \frac{1}{\left| \mathcal{M}_n\right|}\sum_{i\in \mathcal{M}_n} \left\|\mathcal{F}_{v,i}^{'n}-\mathcal{F}_{v,i}\right\|_2^2,
	\label{noise}
\end{align}
where $\mathcal{F}_{v}^{'m}$, $\mathcal{F}_{v}^{'n}$, $\mathcal{F}_v$ denote the reconstructed masked tokens, denoising tokens, and the initial tokens, respectively.

\textbf{Multitask Learning Network.} To more effectively model causal spatio-temporal dependencies between joints based on feature tokens, a naive method is to feed the feature tokens into a self-attention module~\cite{chen2024dualfluidnet}. However, there is a natural spatial association between adjacent joints in a human pose, but the temporal trajectory of each joint is independent~\cite{he2024video}. This indicates the importance of decoupling the spatial and temporal dependencies between joints. Motivated by the observation and insight, we propose a simple but efficient multitask learning network: a criss-cross spatio-temporal attention network, including a spatio-temporal pathway and a temporal-spatio pathway. The primary task and auxiliary tasks share this network, where the auxiliary tasks promote the network to infer unknown tokens based on normal tokens, giving the network causal spatio-temporal modeling capabilities. Concretely, the input of these tasks can be obtained by:
\begin{align}
	X_v = \mathcal{F}_v\oplus \mathcal{F}_k, X_v^m = \mathcal{F}_v^m\oplus \mathcal{F}_k, X_v^n = \mathcal{F}_v^n\oplus \mathcal{F}_k,
\end{align}
where $\oplus$ denotes the concatenate operation, $X_v, X_v^m,$ and $X_v^n$ are the input tokens of three tasks. $\mathcal{F}_v$, $\mathcal{F}_v^m$, $\mathcal{F}_v^n$ and $\mathcal{F}_k$ represent the initial feature tokens, masked tokens, noised tokens, and the keypoint tokens, respectively. The output of multitask learning network is formulated as:

\begin{align}
	\{\mathcal{F}_v^{'};\mathcal{F}_k^{'}\} = D_T(D_S(X_v)) \oplus D_S(D_T(X_v)),\\
	\{\mathcal{F}_v^{'m};\mathcal{F}_k^{'m}\} = D_T(D_S(X_v^m)) \oplus D_S(D_T(X_v^m)),\\
	\{\mathcal{F}_v^{'n};\mathcal{F}_k^{'n}\} = D_T(D_S(X_v^n)) \oplus D_S(D_T(X_v^n)),
\end{align}
where $D_T$ and $D_S$ represent the temporal and spatial self-attention module, respectively. \textsc{$\mathcal{F}_v^{'}$}, $\mathcal{F}_v^{'m}$, and $\mathcal{F}_v^{'n}$ denotes the refined coarse tokens, the reconstructed masked tokens, and the denoising tokens, respectively. $\mathcal{F}_k^{'}$, $\mathcal{F}_k^{'m}$, and $\mathcal{F}_k^{'n}$ represent the corresponding learned keypoint tokens. 

\subsection{Fine Token Enhancement}
\label{Finetoken}
In human pose estimation tasks, causal feature tokens are more valuable than the non-causal tokens (such as background, objects, \textit{etc}.). An intuitive approach is to utilize the causal tokens and discard the other tokens to estimate the human pose. However, motivated by the observation and insight in~\cite{xiao2020noise,long2023beyond}, image background could improve the accuracy of vision task due to their potential and implicit relations to the human. 


\textbf{Tokens Causal Importance Selection.} To prioritize the causal tokens for pose estimation, we propose tokens causal importance selection module. We first employ a temporal fusion layer to aggregate the multi-frame feature tokens $\mathcal{F}_v^{'}$ to a tokens $\mathcal{\hat{F}}_v$. To endow the interpretability of our method, we decouple the tokens $\mathcal{\hat{F}}_v$ into two categories (\textit{i.e.}, the causal and the non-causal tokens) by comparing the similarity with the learned keypoint tokens $\mathcal{F}_k^{'}$. Mathematically, we compute the similarity score between the keypoint tokens and all feature tokens as:
\begin{align}
	S^i = Softmax(\frac{\mathbf{Q}_k^i\mathbf{K}_v^T}{\sqrt{D}}),
\end{align}
where $\mathbf{Q}_k^i$ represents the query vector of the $i^{th}$ keypoint token $\mathcal{F}_k^{'i}$, $\mathbf{K}_v$ is the key matrix of feature tokens $\mathcal{\hat{F}}_v$, $D$ is the latent dimension. For each keypoint token $\mathcal{F}_k^{'i}$, we select the top-$n$ tokens as causal tokens for $i^{th}$ keypoint token according to similarity scores $S^i$. For $K$ keypoint tokens, we can obtain $nK$ causal tokens $\mathcal{\hat{F}}_v^c$ and the remaining $N-nK$ tokens are set as non-causal tokens $\mathcal{\hat{F}}_v^{nc}$.

\textbf{Non-Causal Tokens Clustering.} Although non-causal tokens provide implicit and potential clues, several non-causal tokens often correspond to the identical region (\textit{e.g.}, background, objects) and the semantic information may be redundant. To this end, we propose a non-causal tokens clustering module, which cluster the non-causal tokens $\mathcal{\hat{F}}_v^{nc}$ and then merge the tokens from the identical group into new tokens through importance weight. Specifically, we employ a density peak clustering algorithm based on k-nearest neighbor (DPC-KNN)~\cite{zeng2022not} to cluster feature tokens. DPC-KNN follows two assumptions: (1) the density of the cluster centers is higher than other samples around them and (2) the distance between different cluster centers is far. This derives two concepts: local density $\rho$ and relative distance $\delta$ for each non-causal token. Given non-causal tokens, $\rho$ and $\delta$ are formulated as:
\begin{align}
	\rho_i = {\rm exp}	\left (-\frac{1}{k}\sum_{f_j\in {\rm KNN}(f_i)}\left\|f_i-f_j \right\|_2^2\right )
	\label{rho},\\
	\delta_i = \begin{cases}
		{\rm min}_{j:\rho_j>\rho_i}(\left\|f_i-f_j \right\|_2) & \text{if}\ \exists\ j\ {\rm s.t.}\ \rho_j>\rho_i \\
		{\rm max}_j (\left\|f_i-f_j \right\|_2) & \text{otherwise}
	\end{cases},
\end{align}
where $f_i, f_j \in \mathcal{\hat{F}}_v^{nc}$, and $\mathcal{\hat{F}}_v^{nc}$ represents the non-causal tokens. $\rho_i, \delta_i$ denote the local density and relative distance of token $f_i$, $KNN(f_i)$ denotes a set of k-nearest neighbors of token $f_i$.
We consider two factors (\textit{i.e.}, $\rho_i$ and $\delta_i$) to determine the cluster center score of token $f_i$ as $\rho_i \times \delta_i$. We choose the top-L tokens with the highest scores as cluster centers, and then assign the remaining tokens to the cluster center with the closest distance.

Each cluster may contain a different number of tokens, and different tokens may have different importance. Inspired by~\cite{long2023beyond}, instead of mindlessly averaging the tokens in the same group, we merge the tokens by a weighted sum. Specifically, we introduce the keypoint token attention to denote importance~\cite{scMFC}, we merge the same group of tokens into a new token as:
\begin{align}
	\bar{f}_i = \sum_{j\in C_i}s_j f_j,
\end{align}
where $\bar{f}_i$ denotes the merged token from $C_i$, $C_i$ denotes the $i^{th}$ cluster, $s_j$ represents the importance score of token $f_j$. We concatenate all $\bar{f}_i$ to get merged non-causal tokens $\mathcal{\hat{F}}_v^{nc}$.

\textbf{Heatmap Generation.} Finally, we aggregate causal tokens $\mathcal{\hat{F}}_v^c$ and the merged non-causal tokens $\mathcal{\hat{F}}_v^{nc}$ to obtain a fine feature tokens $\mathbb{F}_v$ as follows:
\begin{align}
	\mathbb{F}_v = \mathcal{\hat{F}}_v^c \oplus \mathcal{\hat{F}}_v^{nc},
\end{align}
where $\oplus$ represents the concatenate operation. $\mathbb{F}_v$ is then fed into a detection head to obtain the pose heatmaps $\mathbb{H}_t$.

\subsection{Loss Functions} We use a heatmap loss to supervise the pose estimation:
\begin{align}
	\mathcal{L}_H = \left\|\mathbb{H}_t-\mathbb{G}_t \right\|_2^2,
\end{align}
where $\mathbb{H}_t$ and $\mathbb{G}_t$ denote the estimated heatmap and the ground truth heatmap, respectively. For auxiliary task learning, we adopt the reconstruction loss $\mathcal{L}_{mask}$ in the Eq.~\ref{mask} and $\mathcal{L}_{denoise}$ in the Eq.~\ref{noise} to supervise the \textit{multitask learning network}, aiming to capture more causal spatio-temporal modeling ability. The total loss is given by:
\begin{align}
	\mathcal{L}_{total} = \mathcal{L}_H+\lambda(\mathcal{L}_{mask}+\mathcal{L}_{denoise}),
	\label{loss}
\end{align}
where $\lambda$ is a weight hyper-parameter.
\section{Experiments}
\subsection{Experimental Setup}
\textbf{Dataset.} We evaluate the proposed CM-Pose for video-based human pose estimation in three widely used datasets: PoseTrack2017 \cite{iqbal2017posetrack}, PoseTrack2018 \cite{andriluka2018posetrack}, and PoseTrack2021 \cite{doering2022posetrack21}. \textbf{PosTrack2017} includes 80,144 pose annotations and has two subsets, \textit{i.e.}, training (train) and validation (val) with 250 videos and 50 videos (split according to the official protocol), respectively. \textbf{PoseTrack2018} largely increases the number of video clips and pose annotations including 593 videos for training, 170 videos for validation, and the total number of pose annotations is 153,615. PoseTrack2018 also introduces an additional flag characterizing joint visibility. \textbf{PoseTrack2021} further increases the number of pose annotations for small or crowded persons, including 177,164 labels. All three datasets identify 15 keypoints and the training set is densely labeled in the center 30 frames, while the validation set contains additional pose annotations every 4 frames.

\textbf{Implementation Details.} We implement our method CM-Pose for human pose estimation with Pytorch, which is trained on 2 Nvidia Geforce RTX 4090 GPUs and terminated with 20 epochs. We use Vision Transformer (ViT) \cite{dosovitskiy2020image}, pre-trained on the COCO dataset, as the backbone network. For data augmentation, we adopt random scale with a factor of $\pm 0.35$, random rotation $[-45^\circ,45^\circ]$, truncation, and flipping. We set the image size as $256 \times 192$. The time span $\omega$ is set to 1. The number of keypoint tokens $K$ is 15. 
We use AdamW optimizer to train the model with an initial learning rate of $2e-4$ (decays to $2e-5, 2e-6, 2e-7$ at the 5-th, 12-th, 18-th epochs, respectively). 

\textbf{Evaluation Metric.} As in previous works \cite{liu2022temporal}, we use average precision (AP) to evaluate the performance of our model. We calculate the AP for each keypoint and average them to get the final performance (mAP).

\begin{table*}[t] 
	\centering
	\fontsize{8pt}{8pt}\selectfont
	\resizebox{0.80\linewidth}{!}{
		\begin{tabular}{l|ccccccc|c}
			\hline
			Method & Head & Shoulder & Elbow & Wrist & Hip & Knee & Ankle & Mean \\
			\hline
			HRNet \cite{sun2019deep}          & 82.1 & 83.6 & 80.4 & 73.3 & 75.5 & 75.3 & 68.5 & 77.3 \\
			CorrTrack \cite{rafi2020self}      & 86.1 & 87.0 & 83.4 & 76.4 & 77.3 & 79.2 & 73.3 & 80.8 \\
			Dynamic-GNN \cite{yang2021learning}    & 88.4 & 88.4 & 82.0 & 74.5 & 79.1 & 78.3 & 73.1 & 81.1 \\
			PoseWarper \cite{bertasius2019learning}     & 81.4 & 88.3 & 83.9 & 78.0 & 82.4 & 80.5 & 73.6 & 81.2 \\
			DCPose \cite{liu2021deep}         & 88.0 & 88.7 & 84.1 & 78.4 & 83.0 & 81.4 & 74.2 & 82.8 \\
			SLT-Pose \cite{gai2023spatiotemporal}  & 88.9 & 89.7 & 85.6 & 79.5 & 84.2 & 83.1 & 75.8 & 84.2 \\
			HANet \cite{jin2023kinematic}  & 90.0 & 90.0 & 85.0 & 78.8 & 83.1 & 82.1 & 77.1 & 84.2 \\
			M-HANet \cite{jin2024masked}  & 90.3 & 90.7 & 85.3 & 79.2 & 83.4 & 82.6 & 77.8 & 84.8 \\
			FAMI-Pose \cite{liu2022temporal}      & 89.6 & 90.1 & 86.3 & 80.0 & 84.6 & 83.4 & 77.0 & 84.8 \\
			DSTA \cite{he2024video}      & 89.3 & 90.6 & 87.3 & 82.6 & 84.5 & 85.1 & 77.8 & 85.6 \\
			TDMI \cite{feng2023mutual}      & \textbf{90.0} & 91.1 & 87.1 & 81.4 & 85.2 & 84.5 & 78.5 & 85.7 \\
			\hline
			\textbf{CM-Pose (Ours)} & 89.2 & \textbf{92.0} & \textbf{89.0} & \textbf{85.6} & \textbf{88.6} & \textbf{87.2} & \textbf{81.1} & \textbf{87.5} \\
			\hline
	\end{tabular}}
	\caption{Quantitative results on the \textbf{PoseTrack2017} dataset.}
	\label{2017}
\end{table*}

\begin{table*}[t] 
	\centering
	\fontsize{8pt}{8pt}\selectfont
	\resizebox{0.80\linewidth}{!}{
		\begin{tabular}{l|ccccccc|c}
			\hline
			Method & Head & Shoulder & Elbow & Wrist & Hip & Knee & Ankle & Mean \\
			\hline
			Dynamic-GNN ~\cite{yang2021learning} & 80.6 & 84.5 & 80.6 & 74.4 & 75.0 & 76.7 & 71.8 & 77.9 \\
			PoseWarper ~\cite{bertasius2019learning} & 79.9 & 86.3 & 82.4 & 77.5 & 79.8 & 78.8 & 73.2 & 79.7 \\
			DCPose~ \cite{liu2021deep} & 84.0 & 86.6 & 82.7 & 78.0 & 80.4 & 79.3 & 73.8 & 80.9 \\
			SLT-Pose ~\cite{gai2023spatiotemporal} & 84.3 & 87.5 & 83.5 & 78.5 & 80.9 & 80.2 & 74.4 & 81.5 \\
			FAMI-Pose ~\cite{liu2022temporal} & 85.5 & 87.7 & 84.2 & 79.2 & 81.4 & 81.1 & 74.9 & 82.2 \\
			HANet ~\cite{jin2023kinematic} & 86.1 & 88.5 & 84.1 & 78.7 & 79.0 & 80.3 & 77.4 & 82.3 \\
			M-HANet ~\cite{jin2024masked} & \textbf{86.7} & \textbf{88.9} & 84.6 & 79.2 & 79.7 & 81.3 & 78.7 & 82.7 \\
			DSTA ~\cite{he2024video} & 85.9 & 88.8 & 85.0 & 81.1 & 81.5 & 83.0 & 77.4 & 83.4 \\
			TDMI ~\cite{feng2023mutual} & 86.2 & 88.7 & 85.4 & 80.6 & 82.4 & 82.1 & 77.5 & 83.5 \\
			\hline
			\textbf{CM-Pose (Ours)} & 85.7 & \textbf{88.9} & \textbf{85.8} & \textbf{81.0} & \textbf{84.4} & \textbf{84.2} & \textbf{80.1} & \textbf{84.4} \\
			\hline
	\end{tabular}}
	\caption{Quantitative results on the \textbf{PoseTrack2018} dataset.}
	\label{2018}
\end{table*}

\subsection{Comparison with State-of-the-art Methods}
\textbf{Results on the PoseTrack2017 Dataset.} We evaluate our method CM-Pose with existing 11 pose estimation methods, and the results are tabulated in Table~\ref{2017}. Compared to previous methods, the proposed CM-Pose achieves a new state-of-the-art performance of 87.5 mAP and delivers a gain of 1.8 mAP over the best-performing previous work TDMI~\cite{feng2023mutual}. Specifically, we also observe that the encouraging improvement for challenging joints (\textit{i.e.}, wrist, hip): with an mAP of 85.6 ($\uparrow$ 3.0) for wrists and an mAP of 88.6 ($\uparrow$ 3.4) for hips. As would be expected, the proposed auxiliary task learning (\textit{i.e.}, masked token reconstruction task, denoising token task) helps the network to infer challenging keypoints from known keypoints based on spatio-temporal dependencies and causal reasoning ability, which is especially important for challenging scenes such as pose occlusion and video defocus. 



\textbf{Results on the PoseTrack2018 Dataset.} We further evaluate our model on the PoseTrack2018 dataset. Table~\ref{2018} reports the empirical comparisons on validation set. We can observe that our method one again surpasses the state-of-the-art methods, reaching an mAP of 84.4, with an mAP of 84.4, 84.2, 80.1 for the hip, knee, and ankle joints.


\textbf{Results on the PoseTrack2021 Dataset.} We present our results of PoseTrack2021 dataset in Table~\ref{2021}, comparing our model with previous state-of-the-art methods. We see that existing methods such as TDMI~\cite{feng2023mutual} and DSTA~\cite{he2024video} have already reached an impressive performance of 83.5 mAP. In contrast, the proposed method CM-Pose achieves a 84.3 mAP. We also obtain an 88.9 ($\uparrow$ 3.1) mAP for the head, 83.7 ($\uparrow$ 1.3) mAP for the knee, and 84.6 ($\uparrow$ 1.1) mAP for hip joints.

\begin{table*}[t] 
	\centering
	\fontsize{8pt}{8pt}\selectfont
	\resizebox{0.80\linewidth}{!}{
		\begin{tabular}{l|ccccccc|c}
			\hline
			Method & Head & Shoulder & Elbow & Wrist & Hip & Knee & Ankle & Mean \\
			\hline
			CorrTrack ~\cite{rafi2020self} & - & - & - & - & - & - & - & 72.3 \\
			CorrTrack* ~\cite{rafi2020self} & - & - & - & - & - & - & - & 72.7 \\
			DCPose ~\cite{liu2021deep} & 83.2 & 84.7 & 82.3 & 78.1 & 80.3 & 79.2 & 73.5 & 80.5 \\
			FAMI-Pose ~\cite{liu2022temporal} & 83.3 & 85.4 & 82.9 & 78.6 & 81.3 & 80.5 & 75.3 & 81.2 \\
			DSTA ~\cite{he2024video} & 87.5 & 87.0 & 84.2 & 81.4 & 82.3 & 82.5 & 77.7 & 83.5 \\
			TDMI ~\cite{feng2023mutual} & 85.8 & 87.5 & \textbf{85.1} & 81.2 & 83.5 & 82.4 & 77.9 & 83.5 \\
			\hline
			\textbf{CM-Pose (Ours)} & \textbf{88.9} & \textbf{88.3} & 84.4 & \textbf{81.9} & \textbf{84.6} & \textbf{83.7} & \textbf{78.8} & \textbf{84.3} \\
			\hline
	\end{tabular}}
	\caption{Quantitative results on the \textbf{Posetrack2021} dataset. CorrTrack* denotes CorrTrack without ReID.}
	\label{2021}
\end{table*}

\begin{figure}[t]
	\centering
	\includegraphics[width=0.46\textwidth]{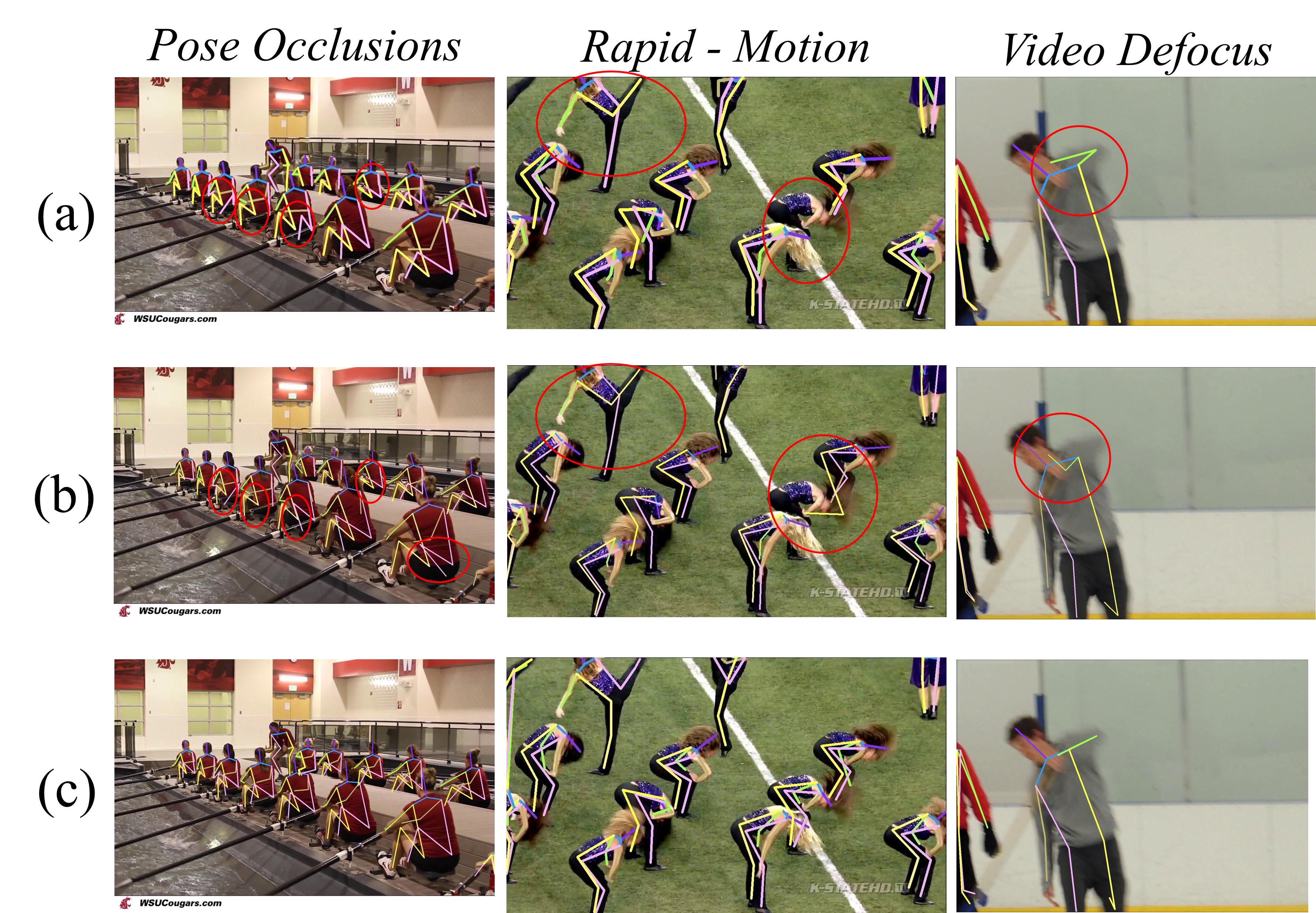}
	\caption{The keyframe (a) and visual comparisons of results obtained from FAMI (a), TDMI (b), and our CM-Pose (c) on challenging scenes in the PoseTrack dataset. Inaccurate predictions are highlighted with the red solid circles.}
	\label{compare}
\end{figure}

\begin{table}[t]
	\centering
	\ra{1.4} 
	\scalebox{0.85}{
		\begin{tabular}{ccc|c|c} 
			\hline
			Primary & Mask & Denoise &Mean &Declines\\
			\hline
			\checkmark &\checkmark&\checkmark & 87.5  & - \\
			\hline
			\checkmark & \checkmark & - & 86.9 &  0.6 ( $\downarrow$) \\
			\checkmark & - & \checkmark & 87.0 &  0.5 ( $\downarrow$)\\
			\checkmark & - & - & 85.6 & 1.9 ( $\downarrow$) \\
			\hline
	\end{tabular}}
	\caption{Ablation study on auxiliary taks in \textbf{MCL}.}
	\label{auxiliary}
\end{table}
\begin{table}[t]
	\centering
	\ra{1.4} 
	\scalebox{0.85}{
		\begin{tabular}{c|cc|c|c} 
			\hline
			Ablation & $\mathcal{\hat{F}}_v^c$ & $\mathcal{\bar{F}}_v^{ir}$ &Mean &Declines\\
			\hline
			CM-Pose & \checkmark &\checkmark& 87.5 & - \\
			\hline
			(a) & \checkmark & - & 85.4 &  2.1( $\downarrow$) \\
			(b)&- & - & 86.0 &  1.5 ( $\downarrow$)\\
			\hline
	\end{tabular}}
	\caption{Ablation of different designs in \textbf{FTE}.}
	\label{token}
\end{table}

\textbf{Comparison of Visual Results.} We further visualize the results with challenging scenarios such as pose occlusion and video defocus to examine the robustness of our method. We illustrate in Figure~\ref{compare} the side-by-side comparisons of state-of-the-art approaches a) FAMI~\cite{liu2022temporal}, b) TDMI~\cite{feng2023mutual} and c) our CM-Pose. From the Figure~\ref{compare}, we see that our CM-Pose consistently estimates more robust and accurate humanpose for challenging scenes. FAMI and TDMI design feature alignment or difference to model spatio-temporal dependencies between joints. Through the integration of auxiliary tasks, our CM-Pose capture more comprehensive spatio-temporal dependencies and grasp better causal reasoning capability. On the other hand, by prioritizing causal tokens and integrating potential information in non-causal tokens, CM-Pose have a better interpretability for reliable pose estimation. 

\begin{figure}[t]
\centering
\includegraphics[width=0.35\textwidth]
{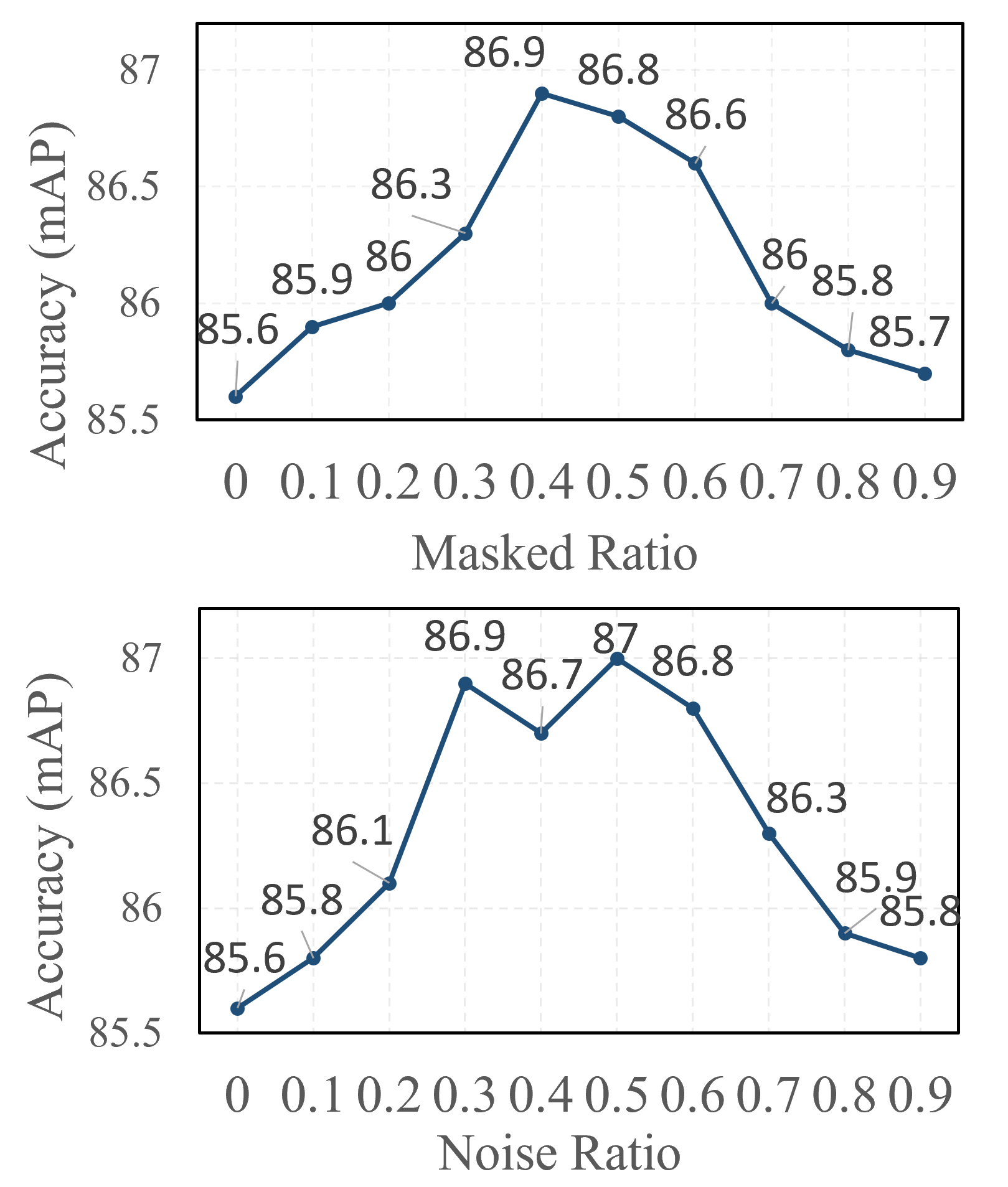}
\caption{Impact of different masked ratios and noise ratios in training process on PoseTrack2017 validation set.}
\label{ratio}
\end{figure}


\subsection{Ablation Study} We perform ablative analysis to examine the contribution of each component in the proposed CM-Pose, including Multitask Coarse Learning (MCL) and Fine Token Enhancement (FTE). We also investigate the effect of different ratio of corrupted tokens (\textit{i.e.}, $p_{tm}$ and $p_{tn}$). These experiments are performed on the PoseTrack2017 validation dataset.

\textbf{Study on Components of MCL.} We explore the effectiveness of the two auxiliary tasks in the MCL: masked token reconstruction task (``Mask'') and denoising token task (``Denoise''), and show the result in Table~\ref{auxiliary}. ``Primary" represents the primary task. From the second, third, and fourth lines of Table~\ref{auxiliary}, we can see that introducing the masked token reconstruction task and denoising token task alone could improve the performance of pose estimation, proving the effectiveness of these auxiliary tasks. Additionally, as shown in the second and fifth line of Table~\ref{auxiliary}, we also find that jointly adopting two auxiliary tasks can further improve the pose estimation performance.

\textbf{Study on the Masked and Noise Ratios.} We perform experiments to explore the effectiveness of different masked ratio $p_{tm}$ and noise ratio $p_{tn}$. As shown in Fig.~\ref{ratio}, we can observe that if the masked ratio $p_{tm}$ and noise ratio $p_{tn}$ are set too low or too high, both lead to poor performance. We conjecture the possible reason is that the inappropriate ratios make the auxiliary task unsuitable for model learning. Furthermore, a moderate ratio makes the model work best, where the appropriate range of masked ratio $p_{tm}$ is 0.4 to 0.5 and noise ratio $p_{tn}$ is 0.3 to 0.6.

\textbf{Study on Components of FTE.} We further verify the impact of the fine token enhancement under different setting and tabulated in Table~\ref{token}. (a) For the first experiment setting, we discard the non-causal tokens $\mathcal{\bar{F}}_v^{nc}$ and only utilize causal tokens $\mathcal{\hat{F}}_v^c$ to estimate the human pose. The mAP performance drops from 87.5 to 85.4. This results deterioration on top of the non-causal tokens also include potentially useful semantic information for pose estimation. (b) For the next experiment setting, we remove the FTE module from CM-Pose. It should be noted that after removing FTE, the tokens from temporal fusion layer are fed to detection head. We see that the performance drops from 87.5 to 86.0 mAP. This indicates the importance of our FTE module, which could pay more attention to causal tokens while also understanding the potential contribution of non-causal tokens to the pose estimation. Interestingly, the setting (b) works better than (a), which also demonstrates the necessity of retaining non-causal tokens.

\vspace{-0.7em}

\section{Conclusion}
In this work, we explore the video-based human pose estimation task from a causal perspective. We propose a novel causal-inspired multitask learning framework, termed CM-Pose, which to the best of our knowledge is the first to leverage auxiliary tasks rather than elaborate network design to grasp more causal spatio-temporal modeling ability. To further identify the causal tokens and integrate non-causal tokens, we present a Token Causal Importance Selection module and a Non-Causal Tokens Clustering module. These modules enhance the interpretability of our model and leads to a more reliable pose estimation results. Empirical experiments on three datasets show our method delivers state-of-the-art performance and equips with higher robustness on challenging scenes. In the future, we will try different auxiliary tasks for human pose estimation and extend our effort to other tasks~\cite{yao2024qe,zhang2024caption}.
\section{Acknowledgments} 
This research is supported by the National Natural Science Foundation of China (No. 62276112, No. 62372402), Key Projects of Science and Technology Development Plan of Jilin Province (No. 20230201088GX), the Key R\&D Program of Zhejiang Province (No. 2023C01217), and Graduate Innovation Fund of Jilin University (No. 2024CX089).
\bibliography{aaai25}

\end{document}